# PARSING OF MYANMAR SENTENCES WITH FUNCTION TAGGING

Win Win Thant[1], Tin Myat Htwe[2] and Ni Lar Thein[3]

[1,3]University of Computer Studies, Yangon, Myanmar
winwinthant@gmail.com
nilarthein@gmail.com
[2]Natural Language Processing Laboratory
University of Computer Studies, Yangon, Myanmar
tinmyathtwe@gmail.com

## ABSTRACT

*This paper describes the use of Naive Bayes to address the task of assigning function tags and context free grammar (CFG) to parse Myanmar sentences. Part of the challenge of statistical function tagging for Myanmar sentences comes from the fact that Myanmar has free-phrase-order and a complex morphological system. Function tagging is a pre-processing step for parsing. In the task of function tagging, we use the functional annotated corpus and tag Myanmar sentences with correct segmentation, POS (part-of-speech) tagging and chunking information. We propose Myanmar grammar rules and apply context free grammar (CFG) to find out the parse tree of function tagged Myanmar sentences. Experiments show that our analysis achieves a good result with parsing of simple sentences and three types of complex sentences.*

## 1. INTRODUCTION

The natural language processing community is in the strong position of having many available approaches to solve some of its most fundamental problems [1]. We have taken Myanmar language for information processing. Myanmar is an agglutinative language with a very productive inflectional system. This means that for any NLP application on Myanmar to be successful, some amount of functional analysis is necessary. Without it, the development of grammatical relations would not be feasible due to the sparse data problem bound to exist in the training data. Our approach is a part of the Myanmar to English machine translation project. If high quality translation is to be achieved, language understanding is a necessity. One problem in Myanmar language processing is the lack of grammatical regularity in the language. This leads to very complex Myanmar grammar in order to obtain satisfactory results, which in term increases the complexity in the parsing process, it is desired that simple grammar is to be used.

Our proposed method makes use of two components. They are function tagging and parsing. Function tags are useful for any application trying to follow the thread of the text –they find the 'who does what' of each clause, which can be useful to gain information about the situation or to learn more about the behaviour of words in the sentence [2]. The goal of function tagging is to assign syntactic categories like subject, object, time and location to each word in the text document. In case of function tagging, we use Naive Bayes theory and the functional annotated tagged corpus. Parsing is the process of analyzing a text or sentence that is made up of a sequence of words called tokens, and to determine its grammatical structure with respect to a given grammatical rules. The goal of the second one is to produce the parse tree of the sentences in Myanmar text.

In our approach, we take the chunk level phrase with the combination of POS tag and its category which is the output of a fully described morphological analyzer [3][4], which is very important for agglutinative languages like Myanmar. A small corpus annotated manually serves as training data because the large scale Myanmar Corpus is unavailable at present. Since the large-scale annotated corpora, such as Penn Treebank, have been built in English, statistical knowledge extracted from them has been shown to be more and more crucial for natural language disambiguation [5]. As a distinctive language, Myanmar has many characteristics different from English. The use of statistical information efficiently in Myanmar language is still a virgin land waiting to explore.

The rest of the paper is organized as in the followings. Next, in the Related Work section, we analyze previous efforts related to the tasks of function tagging and parsing. Section 3 explains Myanmar language. Section 4 describes about corpus statistics. Section 5 explains the procedure of proposed system. Section 6 includes the function tag sets. Section 7 describes about the proposed grammar for Myanmar language. Function tagging model is presented in section 8. Section 9 describes about parsing of Myanmar simple and complex sentences. Section 10 explains about experimental results. Finally the conclusion of the paper is presented.

## 2. RELATED WORK

Blaheta and Johnson [6] addressed the task of function tags assignment. They used a statistical algorithm based on a set of features grouped in trees, rather than chains. The advantage was that features can better contribute to overall performance for cases when several features are sparse. When such features are conditioned in a chain model the sparseness of a feature can have a dilution effect of an ulterior (conditioned) one.

Mihai Lintean and Vasile Rus[7] described the use of two machine learning techniques, naive Bayes and decision trees, to address the task of assigning function tags to nodes in a syntactic parse tree. They used a set of features inspired from Blaheta and Johnson [6]. The set of classes they used in their model corresponds to the set of functional tags in Penn Treebank. To generate the training data, they have considered only nodes with functional tags, ignoring nodes unlabeled with such tags. They trained the classifiers on sections 1-21 from Wall Street Journal (WSJ) part of Penn Treebank and used section 23 to evaluate the generated classifiers.

Yong-uk Park and Hyuk-chul Kwon [8] tried to disambiguate for syntactic analysis system by many dependency rules and segmentation. Segmentation is made during parsing. If two adjacent morphemes have no syntactic relations, their syntactic analyzer makes new segment between these two morphemes, and find out all possible partial parse trees of that segmentation and combine them into complete parse trees. Also they used adjacent-rule and adverb subcategorization to disambiguate of syntactic analysis. Their syntactic analyzer system used morphemes for the basic unit of parsing. They made all possible partial parse trees on each segmentation process, and tried to combine them into complete parse trees.

Mark-Jan Nederhof and Giorgio Satta[9] considered the problem of parsing non-recursive context-free grammars, i.e., context-free grammars that generate finite languages and presented two tabular algorithms for these grammars. They presented their parsing algorithm, based on the CYK (Cocke–Younger–Kasami) algorithm and Earley's alogrithm. As parsing CFG (context-free grammar), they have taken a small hand-written grammar of about 100 rules. They have ordered the input grammars by size, according to the number of nonterminals (or the number of nodes in the forest, following the terminology by Langkilde (2000)).

Kyongho Min and William H. Wilson [10] discussed the robustness of four efficient syntactic error-correcting parsing algorithms that are based on chart parsing with a context-free grammar. They implemented four versions of a bottom-up error-correcting chart parser: a basic bottom-up chart parser, and chart parsers employing selectivity, top-down filtering, and a combination of selectivity and a top-down filtering. They detected and corrected syntactic errors using a system component called IFSCP (Ill-Formed Sentence Chart Parser) described by Min & Wilson (1994), together with a spelling correction module. They tested 4 different lengths of sentences (3, 5, 7, and 11) and 5 different error types, with a grammar of 210 context-free rules designed to parse a simple declarative sentence with no conjunctions, passivisation, or relative clauses.

## 3. MYANMAR LANGUAGE

Myanmar (formerly known as Burma) is one of the South-East Asian countries. There are 135 ethnic groups living in Myanmar. These ethnic groups speak more than one language and use different scripts to present their respective languages. There are a total of 109 languages spoken by the people living in Myanmar [11]. The Myanmar language is the official language and is more than one thousand years old.

### 3.1. Features of Myanmar Language

Generally Myanmar sentence follows the subject, object, and verb pattern. However the interchange of subject, object is acceptable. Unlike English language Myanmar is syntax of relatively free-phrase-order language. Myanmar phrases can be written in any order as long as the verb phrase is at the end of sentence. This can be easily illustrated with the example "သူသည် စာအုပ်ကို စားပွဲပေါ်တွင် ထားသည်။" (He places the book on the table) as shown in table 1. All are valid sentences [12].

Table 1. Word order in Myanmar language

| Case | Myanmar Sentences | Word order |
|---|---|---|
| Case 1 | သူ စာအုပ်ကို စားပွဲပေါ်တွင် ထားသည်။ | (Subj-Obj-Pla-Verb) |
| Case 2 | သူ စားပွဲပေါ်တွင် စာအုပ်ကို ထားသည်။ | (Subj-Pla-Obj-Verb) |
| Case 3 | စာအုပ်ကို စားပွဲပေါ်တွင် သူ ထားသည်။ | (Obj-Pla-Subj-Verb) |
| Case 4 | စာအုပ်ကို သူ စားပွဲပေါ်တွင် ထားသည်။ | (Obj-Subj-Pla-Verb) |
| Case 5 | စားပွဲပေါ်တွင် သူ စာအုပ်ကို ထားသည်။ | (Pla-Subj-Obj-Verb) |
| Case 6 | စားပွဲပေါ်တွင် စာအုပ်ကို သူ ထားသည်။ | (Pla-Obj-Subj-Verb) |

In all the cases, subject is သူ (He), object is စာအုပ်ကို (the book), place is စားပွဲပေါ်တွင် (on the table) and verb is ထားသည် (places). From the above example, it is clear that phrase order does not determine the functional structure in Myanmar language and permits scrambling. Myanmar language follows Subject-Object-Verb orders in contradiction with English language.

### 3.2. Issues of Myanmar Language

The highly agglutinative language like Myanmar, nouns and verbs get inflected. Many times we need to depend on syntactic function or context to decide upon whether the particular word is a noun or adjective or adverb or post position [12]. This leads to the complexity in Myanmar grammatical relations. A noun may be categorized as common, proper or compound. Similarly, verb may be finite, infinite, gerund or contingent.

A number of issues are affecting the function tagging for Myanmar language.

- The subject or object of the sentence can be skipped, and still be a valid sentence.
  For example:
  ရန်ကုန် - သို့ - သွားသည်။
  Yangon - to - go
  (Go to Yangon)

- Myanmar language makes prominent usage of particles, which are untranslatable words that are suffixed or prefixed to words to indicate level of respect, grammatical tense, or mood.
  For example:
  မောင်မောင် - **များ** - ပထမ - ဆု - ရ - လျှင် - သူ့မိဘများ - က - အံ့သြ - လိမ့်မည်။
  Mg Mg - **particle** - first - prize - wins - if - his parents - PPM - surprise - will
  (If Mg Mg wins the first prize, his parents will surprise.)
- In Myanmar language, an adjective can specialize before or after a noun unlike other languages.
  For example:
  သူသည် - **ချမ်းသာသော** - လူ - တစ်ယောက် - ဖြစ်သည်။
  He - rich - man - a - is
  (or)
  သူသည် - လူ - **ချမ်းသာ** - တစ်ယောက် - ဖြစ်သည်။
  He - man - rich - a - is
  (He is a rich man.)
- The subject /object can be another sentence, which does not contain subject or object.
  For example:
  ကလေးများသစ်ပင်အောက်တွင်ကစားနေသည် ကို ကျွန်တော်မြင်သည်။
  (I see the children playing under the tree.)
- The postpositions of subject phrases or object phrases can be hidden.
  For example:
  သူ**သည်**- ဆရာဝန် -တစ်ယောက် - ဖြစ်သည်။
  He - doctor - a - is
  (or)
  သူ - ဆရာဝန် - တစ်ယောက် - ဖြစ်သည်။
  He - doctor - a - is
  (He is a doctor.)
- The postpositions of time phrases or place phrases can be omitted.
  For example:
  သူမ - ကျောင်း - **သို့** - သွားသည်။
  She - school - **to** - goes
  (or)
  သူမ - ကျောင်း - သွားသည်။
  She - school - goes
  (She goes to school.)

These issues will cause a lot of problem during function tagging, and a lot of possible tags will be resulted.

### 3.3. Grammar of Myanmar Language

Grammar studies the rules behind languages. The aspect of grammar that does not concern meaning directly is called *syntax*. Myanmar (syntax: SOV), because of its use of postposition (wi.Bat), would probably be defined as a "postpositional language", whereas English (syntax: SVO) because of its use of preposition would probably be defined as a "prepositional language".

There are really only two parts of speech in Myanmar, the noun and the verb, instead of the usually accepted eight parts (Pe Maung Tin 1956:195). Most Myanmar linguists [13] accepted there are eight parts of speech in Myanmar. Myanmar nouns and verbs need the help of suffixes or particles to show grammatical relations.

For example:
ကျောင်းသူများ**သာ** ဂုဏ်ထူးရသည်။
သူတို့သည် အတန်းထဲမှာ ရှိ**ကြ**၏။

Myanmar is a highly verb-prominent language and that suppression of the subject and omission of personal pronouns in connected text result in a reduced role of nominals. This observation misses the critical role of postposition particles marking sentential arguments and also of the verb itself being so marked. The key to the view of Myanmar being structures by nominals is found in the role of the particles. Some particles modify the word's part of speech. Among the most prominent of these is the particle အ, which is prefixed to verbs and adjectives to form nouns or adverbs.There is a wide variety of particles in Myanmar [14].

For example:
သူတို့သည် မန္တလေးတွင် ၈ ရက် **တိတိ** လည်ခဲ့သည်။

Stewart remarked that "The Grammar of Burmese is almost entirely a matter of the correct use of particles"(Stewart 1956: xi). How one understands the role of the particles is probably a matter of one's purpose.

### 3.4. Syntacic Structure of Myanmar Language

It is known that many postpositions can be used in a Myanmar sentence. If the words can be misplaced in a sentence, the sentence can be abnormal. There are two kinds of sentence as a sentence construction. They are simple sentence (SS) and complex sentence (CS). In simple sentence, other phrases such as object, time, and place can be added between subject and verb. There are two kinds of clause in a complex sentence called independent clause(IC) and dependent clause (DC).There must be at least one independent clause in a sentence. But there can be more than one dependent clause in it. IC contains sentence's final particle (sfp) at the end of a sentence [15].

SS=IC+sfp
CS=DC...+IC+sfp

IC may be noun phrase or verb or combination of both.

| | |
|---|---|
| IC=N... | (မျက်မှန်နှင့်ကျောင်းသား) |
| IC=V | (စား) |
| IC=N...+V | (ဘုရားမှာပန်းနဲ့ဆီမီးလှူ) |

DC is the same as IC but it must contain a clause marker (cm) in the end.

| | |
|---|---|
| DC=N...+cm | (ကျောင်းကဆရာ+ပဲ) |
| DC=V+cm | (ရောက်+ရင်) |
| DC=N...+V+cm | (စိတ်ထား+ဖြူ+မှ) |

## 4. CORPUS STATISTICS

Corpus is a large and structured set of texts. It is used to do statistical analysis, checking occurrences or validating linguistic rules on a specific universe. Besides, it is a fundamental basis of many researches in Natural Language Processing (NLP). Building of the corpus will be helpful for development NLP tools (such as grammar rules, spelling checking, etc). However, there are very few creations and researches of corpora in Myanmar, comparing to other language such as English.

We collected several types of Myanmar texts to construct a corpus. Our corpus is to be built manually. We extended the POS tagged corpus that is proposed in [3]. The chunk and function tags are manually added to the POS tagged corpus. The number of sentences is about 3900 sentences with average word length 15 and it is not a balanced corpus that is a bit biased on Myanmar textbooks of middle school. The corpus size is bigger and bigger because the tested sentences are automatically added to the corpus. In table 2, Myanmar grammar books and websites are text collections. Example corpus sentence is shown in figure 1.

Table 2. Corpus statistics

| Text types | # of sentences |
|---|---|
| Myanmar textbooks of middle school | 1250 |
| Myanmar Grammar books | 628 |
| Myanmar Newspapers | 730 |
| Myanmar websites | 970 |
| Others | 325 |
| Total | 3903 |

VC@Active[မိုးရွာ/verb.common]#CC@CCS[လျှင်/cc.sent]#NC@Subj[ကလေး/n.person,များ/part.number]#NC@PPla[လမ်း/n.location]#PPC@PlaP[ပေါ်တွင်/ppm.place]#NC@Obj[ဘောလုံး/n.objects]#VC@Active [ကန်ကြ/verb.common]#SFC@Null[သည်/sf]။

Figure 1. A sentence in the corpus

## 5. PROPOSED SYSTEM

The procedure of the proposed approach is shown in the following figure.

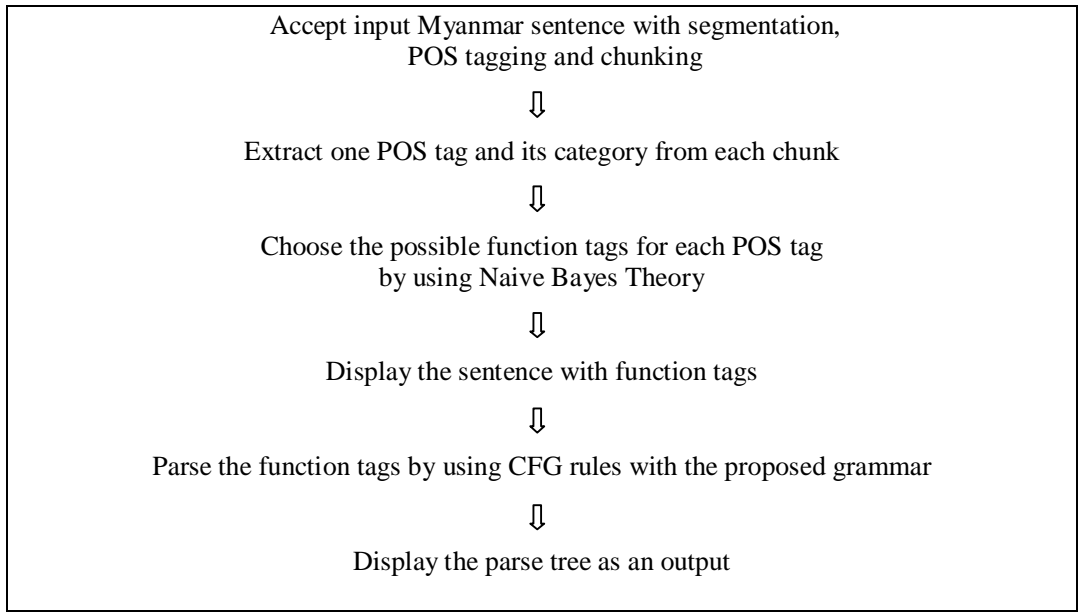

Figure 2. Proposed system

## 6. FUNCTION TAGSET

Function tagging is a process of assigning syntactic categories like subject, object, time and location to each word in the text document. These are conceptually appealing by encoding an event in the format of "who did what to whom, where, when", which provides useful semantic information of the sentences. We use the function tags that is proposed in [16] because it is easier to maintain and can add new language features. The function tagset is shown in table 3.

Table 3. Function tagset

| Tag | Description | Example |
| --- | --- | --- |
| Active | Verb | စားသည် |
| Subj | Subject | သူ |
| PSubj | Subject | သူ |
| SubjP | Postposition of Subject | သည် |
| Obj | Object | ကော်ဖီ |
| PObj | Object | ကော်ဖီ |
| ObjP | Postposition of Object | ကို |
| PIobj | Indirect Object | မလှ |
| IobjP | Postposition of Indirect Object | အား |
| Pla | Place | ရန်ကုန် |
| PPla | Place | ရန်ကုန် |
| PlaP | Postposition of Place | သို့ |
| Tim | Time | မနက် |
| PTim | Time | မနက် |
| TimP | Postposition of Time | တွင် |
| PExt | Extract | ကျောင်းသားများ |
| ExtP | Postposition of Extract | အနက် |
| PSim | Similie | မင်းသမီး |
| SimP | Postposition of Similie | ကဲ့သို့ |
| PCom | Compare | သူဦးလေး |
| ComP | Postposition of Compare | နှင့်အတူ |
| POwn | Own | သူ |
| OwnP | Postposition of Own | ၏ |
| Ada | Adjective | လှ |
| PcomplS | Subject Complement | သူသည်**ဆရာ**ဖြစ်သည် |
| PcomplP | Object Complement | ရွှေကို**လက်စွပ်**လုပ် သည် |
| PPcomplO | Object Complement | ထွန်းထွန်း |
| PcomplOP | Postposition of Object Complement | ဟု |
| PUse | Use | တုတ် |
| UseP | Postposition of Use | ဖြင့် |
| PCau | Cause | မိုး |
| CauP | Postposition of Cause | ကြောင့် |
| PAim | Aim | အမေ့ |
| AimP | Postposition of Aim | အတွက် |
| CCS | Join the sentences | လျှင် |
| CCM | Join the meanings | ထို့ကြောင့် |
| CCC | Join the words | နှင့် |
| CCP | Join with particles | ကို |
| CCA | Join as an adjective | မည့် |

## 7. PROPOSED GRAMMAR FOR MYANMAR SENTENCES

Since it is impossible to cover all types of sentences in Myanmar language, we have taken some portion of the sentence and try to make grammar for them. Myanmar is free-phrase-order language. In Myanmar language, we see that one sentence can be written in different forms for the same meaning, i.e. the positions of the tags are not fixed. So we cannot restrict the grammar rule for one sentence. The grammar rule may be very long, but we have to accept it. The grammar rule we have tried to make, may not work for all the sentences in Myanmar language because we have not considered all types of sentences. Some of the sentences are shown below, which are used to make the grammar rules.

| | |
|---|---|
| သူ-သည်-ကျောင်း-သို့-သွား-သည်။ | (Subj-Pla-Verb) |
| သူ-သည်-ကျောင်းသားတစ်ယောက်-ဖြစ်-သည်။ | (Subj-PcomplS-Verb) |
| ကောင်စီဝင်-အဖြစ်-သူ့-ကို-လူထု-က-ရွေး-သည်။ | (PcomplO-Obj-Subj-Verb) |
| မောင်လှ-သည်-ခွေး-ကို-တုတ်-ဖြင့်-ရိုက်-သည်။ | (Subj-Obj-Use-Verb) |
| သူ-သည်-ဆရာ့-ကို-စာအုပ်-ပေး-သည်။ | (Subj-Obj-Iobj-Verb) |
| သူမ-သည်-လူနာများ-ကို-ဆွေမျိုးများ-ကဲ့သို့-ပြုစု-သည်။ | (Subj-Obj-Sim-Verb) |
| ကလေးများ-သည်-အဖော်-ကြောင့်-ပျက်စီး-သည်။ | (Subj-Cau-Verb) |
| သစ်ရွက်တို့-သည်-တပေါင်းလ-၌-ကြွေ-သည်။ | (Subj-Tim-Verb) |
| တရားသူကြီး-သည်-ခိုးမှု-ကို-တရားရုံး-၌-နံနက်-က-စစ်ဆေး-သည်။ | (Subj-Obj-Pla-Tim-Verb) |
| အမေသည်-သူ့သားအတွက်-မုန့်ကို-ဈေးမှ-မနက်က-ဝယ်ခဲ့သည်။ | (Subj-Aim-Obj-Pla-Tim-Verb) |

Our proposed grammar for Myanmar Sentences:

| | |
|---|---|
| Sentence | →I-sent \| I-sent CC I-sent \| CCM I-sent \| Obj-sent I-sent \| Subj-sent I-sent |
| I-sent | →Subj Obj Pla Active \| Subj Active \| Com Pla Active \| Subj PcomplS Active |
| CC | →CCS \| CCP |
| Subj -sent | →I-sent CCA Subj |
| Obj -sent | →I-sent CCA Obj |
| Subj | →PSubj SubjP |
| Subj | →Subj |
| Obj | →PObj ObjP |
| Obj | →Obj |
| Pla | →PPla PlaP |
| PcomplO | →PPcomplO PcomplOP |
| Use | →PUse UseP |
| Sim | →PSim SimP |

## 8. FUNCTION TAGGING

### 8.1 Naive Bayes Classifier

Before one can build naive Bayesian based classifier, one needs to collect training data. The training data is a set of problem instances. Each instance consists of values for each of the defined features of the underlying model and the corresponding class, i.e. function tag in our case. The development of a Naive Bayes classifier involves learning how much each function tag should be trusted for the decisions it makes [17]. It is well-matched to the function tagging problem.

The Naïve Bayesian classifier is a term in Bayesian statistics dealing with a simple probabilistic classifier based on applying Bayes' theorem with strong (naïve) independence assumptions. It assumes independence among input features. Therefore, given an input vector, its target class can be found by choosing the one with the highest posterior probability. The probability model for a classifier is a conditional model.

$$P(c_k|x_1, x_2, \ldots, x_i) = P(c_k) * P(x_1, x_2, \ldots, x_i | c_k) \qquad (1)$$

Let     $X = x_1, x_2, x_3, \ldots$ ($x_i$, $i >= 1$ and X are features)
        $C = c_1, c_2, c_3, \ldots$ ($c_k$, $k >= 1$ and C are classes)
        $P(c_k|x_1, x_2, \ldots, x_i)$ is referred to as the posterior probability
        $P(c_k)$ as the prior probability
        $P(x_1, x_2, \ldots, x_i|c_k)$ as the log likelihood

## 8.2. Function Tagging by Using Naïve Bayes Theory

The labels such as subject, object, time, etc. are named as function tags. By function, it is meant that action or state which a sentence describes. The system operates at word-level with the assumption that input sentences are pre-segmented, pos-tagged and chunked.

Each proposed function tag is regarded as a class and the task is to find what class/tag a given word in a sentence belongs to a set of predefined classes/tags. A feature is a POS tag word with category. The category of a word is added to the POS tag to obtain more accurate lexical information. It can be formed from the features of that word.

For example: Ma Ma is a clever student.

**Ma Ma** [ မမ(n.person) သည်(ppm.subj) ] **clever** [ စာတော်သော(adj.dem) ] **student** [ကျောင်းသူ(n.person)] **a** [ တစ်(part.number) ယောက်(part.type) ] **is** [ ဖြစ်(v.common) သည် (sf.declarative) ]

Noun has 16 categories such as animals, person, objects, food, location, etc. There are 47 categories in our corpus. We show some features of Myanmar words as shown in table 4.

Table 4. Features

| Feature | English | Myanmar |
|---|---|---|
| n.food | apple | ပန်းသီး |
| pron.possessive | his | သူ့ |
| ppm.time | at | တွင် |
| adj.dem | happy | ပျော်ရွှင်သော |
| part.support | can | နိုင် |
| cc.mean | so | ထို့ကြောင့် |
| v.common | go | သွား |
| sf.declarative | null | ၏ |

In Myanmar language, some words have same meaning but in different features as shown in table 5. For example:

- Ma Ma **and** Hla Hla are friends.
- He lives **with** his uncle.
- He hits the dog **with** the stick.

In these three sentences, English words (and, with, with) have the same Myanmar meaning (နှင့်).

Table 5. Same word with different features

| Feature | English | Myanmar |
|---|---|---|
| cc.chunk | and | နှင့် |
| ppm.compare | with | နှင့် |
| ppm.use | with | နှင့် |

A class is a one of the proposed function tags. Same word may have different function tags as shown in table 6.

Table 6. Function tags

| Function tags | English | Myanmar |
|---|---|---|
| PcomplS | He has a **house**. | အိမ် |
| PPla | He lives in a **house**. | အိမ် |
| PSubj | A **house** is near the school. | အိမ် |
| PObj | He buys a **house**. | အိမ် |

There are many chunks in a sentence such as NC (noun chunk), PPC (postpositional chunk), AC (adjectival chunk), RC (adverbial chunk), CC (conjunctional chunk), SFC (sentence's final chunk) and VC (verb chunk). The chunk types are shown in table 7.

Table 7. Chunk types

| No. | Chunk Type | English | Example |
|---|---|---|---|
| 1 | Noun Chunk | they | NC[သူတို့/pron.person] |
| 2 | Postpositional Chunk | at | PPC[တွင်/ppm.place] |
| 3 | Adjectival Chunk | brave | AC[ရဲရင့်/adj.dem] |
| 4 | Adverbial Chunk | quickly | RC[လျင်မြန်စွာ/adv.manner] |
| 5 | Conjunctional Chunk | or | CC[သို့မဟုတ်/cc.chunk] |
| 6 | Sentence Final Chunk | - | SFC[၏/sf.declarative] |
| 7 | Verb Chunk | help | VC[ကူညီ/v.common] |

A chunk contains a Myanmar head word and its modifier. It can contain more than one POS tag and one of the POS tags is selected with respect to the chunk type. In the following chunk, the POS tag (n.animals) is selected with respect to the chunk type (NC).

For example:   NC [ခွေး/n.animals,တစ်/part.number,ကောင်/part.type]

If the noun chunk (NC) contains more than one noun, the last noun (n.food) is selected as a main word according to the nature of Myanmar language.

For example:   NC [ဆောင်းရာသီ/n.time,သီးနှံပင်/n.food,များ/part.number]

There are many possible function tags ($t_1, t_2…t_k$) for each POS tag with category (pc). These possible tags are retrieved from the training corpus by using the following equation that is prior probability as shown in Table 8.

$$P(t_k|pc) = C(t_k,pc)/C(pc) \qquad (2)$$

Table 8. Sample data for POS/function tag pairs with probability

| POS tags | Function tags : Probability |
|---|---|
| ppm.use | UseP:1.0 |
| n.natural | PSubj:0.209, Subj:0.2985, PPla:0.1343, PObj:0.1642, PcomplS:0.0448, PPcomplO:0.0149, PCau:0.0448, PSim:0.0149, PAim:0.0299, |

|  | Obj:0.0299, PCom:0.0149 |
|---|---|
| pron.possessive | PIobj:0.1111, PSubj:0.2222, PObj:0.6667 |
| cc.chunk | CCC:1.0 |
| adj.dem | PcomplS:0.0192, Ada:0.9808 |
| n.animal | Subj:0.1212, PObj:0.3333, PcomplS:0.1212, PSubj:0.2727, PSim:0.0606, Obj:0.0303, PAim:0.0303, PUse:0.0303 |
| v.common | Active:1.0 |
| part.eg | PcomplOP:0.5455, SimP:0.4545 |

We calculate the probability between next function tags ($n_1, n_2…n_j$) and previous possible tags by using the following equation that is log likelihood as shown in Table 9.

$$P(n_j|t_k) = C(n_j,t_k)/C(t_k) \quad (3)$$

Table 9. Sample data for function/function tag pairs with probability

| Function tags | Function tags : Probability |
|---|---|
| CCC | Subj:0.271, Active:0.2452, PObj:0.1226, Obj:0.129, PTim:0.0194 PcomplS:0.0516, PPla:0.0516, Pla:0.0387, Tim:0.0194, PSubj:0.0387 PCau:0.0065, PAim:0.0065 |
| Subj | CCC:0.2047, Active:0.5436, PTim:0.0067, PCom:0.0067, Ada:0.0604, PDir:0.0067, Tim:0.0134, Pla:0.0101, PUse:0.0034, PSim:0.0101, PLea:0.0134, CCA:0.0268, Obj:0.0503, PPla:0.0235, PObj:0.0168 CCS:0.0034 |
| PCau | CCC:0.1111, CauP:0.8889 |
| PExt | ExtP:1.0 |
| UseP | Active:0.5652, PObj:0.087, Subj:0.087, PArr:0.0435, PTim:0.087, CCA:0.0435, PcomplS:0.0435, Obj:0.0435 |
| PPla | CCC:0.056, PlaP:0.936, PPla:0.0080 |
| Obj | CCC:0.2667, Active:0.6917, AimP:0.0083, Subj:0.0083, CCA:0.0083 Ada:0.0167 |
| PcomplO | Active:1.0 |

Possible function tags are disambiguated by using Naïve Bayesian method. We multiply the probabilities from (2) and (3) and choose the function tag with the largest number as the posterior probability.

Technically, the task of function tags assignment is to generate a sentence that has correct function tags attached to certain words.

Our description of the function tagging process refers to the example as shown in figure 3, which illustrates the sentence ("မမနှင့်လှလှသည် ကျောင်းသို့ စက်ဘီးဖြင့် သွားသည်။" (Ma Ma and Hla Hla go to school by bicycle). This sentence is represented as a sequence of word-tags as "noun verb conjunction noun ppm pronoun verb". It is described as a sequence of chunk as "NC VC CC NC PPC NC VC SFC".

(a)   NC[မမ/n.person]#CC[နှင့်/cc.chunk]#NC[လှလှ/n.person]#PPC[သည်/ppm.subj]#NC[ကျောင်း/n.location] #PPC[သို့/ppm.place]#NC[စက်ဘီး/n.objects]#PPC[ဖြင့်/ppm.use]#VC[သွား/v.common]#SFC[သည်/sf]။

(b)   PSubj[မမ]#CCC[နှင့်]#PSubj[လှလှ]#SubjP[သည်]#PPla[ကျောင်း]#PlaP[သို့]#PUse[စက်ဘီး]#UseP[ဖြင့်] #Active[သွားသည်]။

Figure 3. An overview of function tagging of the sentence
(a)The input POS-tagged and chunk sentence (b) The output sentence with function tags

# 9. Parsing

## 9.1. Context Free Grammar for Myanmar Sentences

The LANGUAGE defined by a CFG (context-free grammar) is the set of strings derivable from the start symbol S (for Sentence). The core of a CFG grammar is a set of production rules that replaces single variables with strings of variables and symbols. The grammar generates all strings that, starting with a special start variable, can be obtained by applying the production rules until no variables remain. A CFG is usually thought in two ways: a device for generating sentences, or a device if assigning a structure to a given sentence. We use CFG for grammatical relations of function tags.

A CFG is a 4-tuple <N,Σ,P,S> consisting of
- A set of non-terminal symbols N
- A set of terminal symbols Σ
- A set of productions P
  - A-> α
  - A is a non-terminal
  - α is a string of symbols from the infinite set of strings (ΣU N)*
- A designated start symbol S

## 9.2. Parsing Simple Sentences

A simple sentence contains one subject and one verb. We can construct simple sentences in many different forms.

- Constructed by adding adjective and adverb
  Adjective +   Subject    + Adjective + Object  + Adverb  + Verb
  ဝသော      +ကောင်လေးသည် +  ချိုသော  + ကိတ်မုန့်ကို + လျင်မြန်စွာ + စားသည်။
  Fat       +    boy     +  sweet    +  cake    + quickly +eat
  (A fat boy eats quickly the sweet cake.)

- Constructed by using different set of phrases
  Subject phrase +    Object phrase   + Verb
  ဦးဘ၏သားသည်  + ဦးထုပ်အနီနှင့်ကောင်လေးကို +ရှာသည်။
  U Ba's son   +   boy with the red hat  + find
  (U Ba's son finds a boy with the red hat.)

- Constructed by omitting subject
  Object +    Time    + Verb
  ဆံပင်ကို +တနင်္ဂနွေနေ့တွင်+လျှော်သည်။
  Hair   + in Sunday + wash
  (Wash the hair in Sunday.)

- Constructed by omitting verb
  Subject + Subject's complement+ Sentence's final particle
  သူက    +       ဆရာ        +ပါ။
  He     +     teacher       + null
  (He is a teacher.)

Consider a simple declarative sentence "သူတို့သည် မောင်ဘကို ခေါင်းဆောင် အဖြစ် ရွေးချယ်ခဲ့ သည်။" (They selected Mg Ba as a leader).

The structure of the above sentence is Subj-Obj-PcomplO-Active. This is a correct sentence according to the Myanmar literature.

(a)  NC[သူတို့/pron.possessive]#PPC[သည်/ppm.subj]#NC[မောင်ဘ/n.person]#PPC[ကို/ppm.obj]#NC[ခေါင်းဆောင်/n.person]#PPC[အဖြစ်/part.eg]#VC[ရွေးချယ်/v.common,ခဲ့/part.support]#SFC[သည်/sf]။

(b)  PSubj[သူတို့]#SubjP[သည်]#PObj[မောင်ဘ]#ObjP[ကို]#PPcomplO[ခေါင်းဆောင် ]#PcomplOP[အဖြစ်]# Active[ရွေးချယ်ခဲ့သည်]။

(c)

| | |
|---|---|
| <u>Sentence</u> | [start] |
| <u>I-sent</u> | [Sentence→I-sent] |
| <u>Subj</u> Obj PcomplO Active | [I-sent→ Subj Obj PcomplO Active] |
| PSubj SubjP <u>Obj</u> PcomplO Active | [Subj → PSubj SubjP] |
| PSubj SubjP PObj ObjP <u>PcomplO</u> Active | [Obj → PObj ObjP] |
| PSubj SubjP PObj ObjP PPcomplO PcomplOP Active | [PcomplO→PPcomplO PcomplOP ] |

(d)

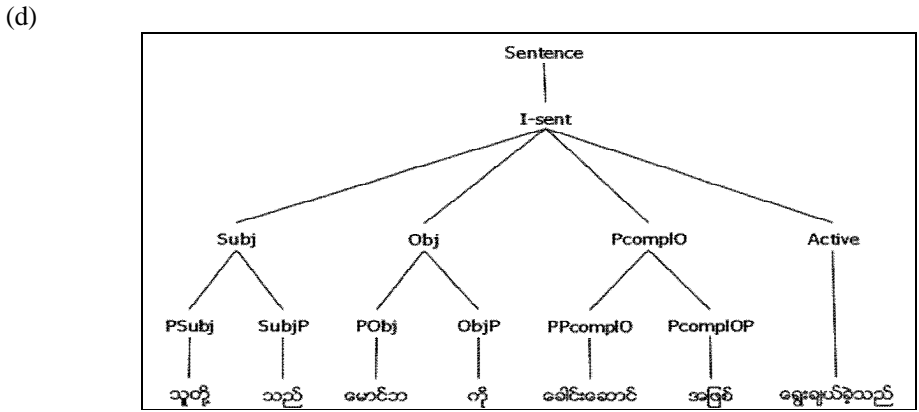

Figure 4. (a) The tagged and chunk simple sentence (b) The function tagged sentence
(c) Grammar derivation for simple sentence (d) The parse tree with function tags

## 9.3. Parsing Complex Sentences

Complex sentence has more than one verb. It contains at least two simple sentences. Simple sentences are joined with postpositions, particles or conjunctions. There are three types of complex sentences.

### 9.3.1. Two simple sentences are joined with postpositions

Consider a complex sentence "သူရေကူးနေသည် ကို ကျွန်တော် တွေ့သည်။" (I see that he is swimming).

In this sentence, two simple sentence သူရေကူးနေသည် (he is swimming) and ကျွန်တော် တွေ့သည် (I see) is joined by postposition ကို (that). The structure of the above sentence is Subj-Active-CCP-Subj-Active. This is a correct sentence according to the Myanmar literature.

(a)  NC [သူ/pron.person] # VC [ရေကူးနေသည်/v.common] # CC [ကို/cc.obj] # NC [ကျွန်တော်/pron.person] # VC [တွေ့/v.common] # SFC [သည်/sf]။

(b)  Subj[သူ] # Active[ရေကူးနေသည်] # CCP[ကို] # Subj[ကျွန်တော်] # Active[တွေ့သည်]။

(c)

| | |
|---|---|
| <u>Sentence</u> | [start] |
| <u>I-sent</u> CCP I-sent | [Sentence→I-sent CCP I-sent] |
| Subj Active CCP <u>I-sent</u> | [I-sent→ Subj Active] |
| Subj Active CCP <u>Subj</u> Active | [I-sent→Subj Active] |

(d)

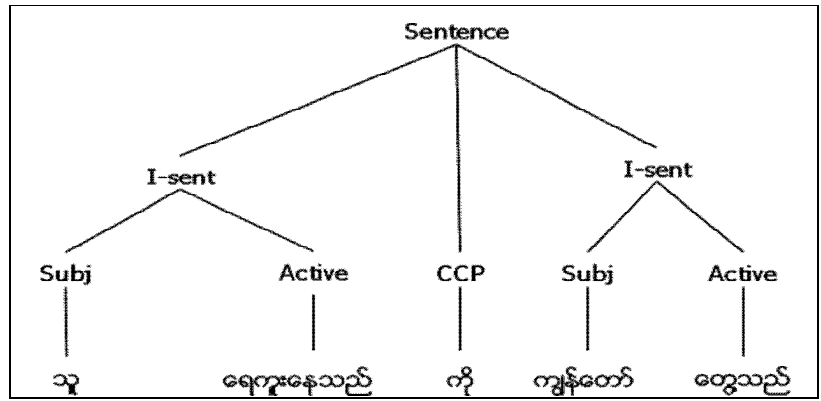

Figure 5. (a) The tagged and chunk complex sentence joined with postposition (CCP)
(b) The function tagged sentence (c) Grammar derivation (d) The parse tree with function tags

### 9.3.2. Two simple sentences are joined with particles

In figure 7, the sentence "အဖေပေးသောစာအုပ်သည် ကောင်းသည်။" (The book that is given by my father is good.) is illustrated. It is described as a sequence of chunk as "NC VC CC NC PPC AC SFC" and the sentence structure (Sentence) contains separate constituents for the subject sentence (Subj-sent) and independent sentence (I-sent), which contains other phrases.

(a)  NC [အဖေ/n.person] # VC [ပေး/v.common] # CC [သော/cc.adj] # NC [စာအုပ်/n.objects] # PPC [သည်/ppm.subj] # AC [ကောင်း/adj.dem] # SFC [သည်/sf]။

(b)  Subj[အဖေ]#Active[ပေး]#CCA[သော]#PObj[စာအုပ်]#ObjP[သည်]#Active[ကောင်းသည်]။

(c)

| Sentence | [start] |
| --- | --- |
| Subj-sent I-sent | [Sentence→Subj-sent I-sent] |
| I-Sent CCA Subj I-sent | [Subj-sent→ I-Sent CCA Subj] |
| Subj Active CCA Subj I-sent | [I-sent→Subj Active] |
| Subj Active CCA PSubj SubjP I-sent | [Subj → PSubj SubjP] |
| Subj Active CCA PSubj SubjP Ada | [I-sent → Ada ] |

(d)

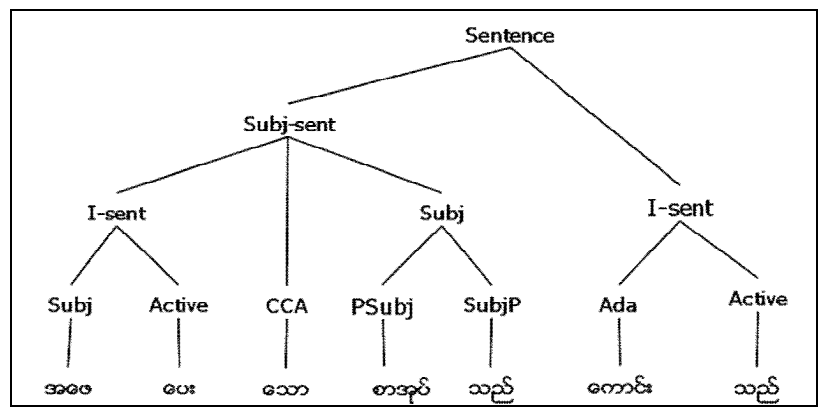

Figure 6. (a) The tagged and chunk complex sentence joined with particle (CCA)
(b) The function tagged sentence (c) Grammar derivation (d) The parse tree with function tags

### 9.3.3. Two simple sentences are joined with conjunctions

Consider a complex sentence "သူလိမ္မာ သောကြောင့် ဆရာများက သူ့ကို ချစ်ကြသည်။" (As he is clever, the teachers love him).

In this sentence, two simple sentence သူလိမ္မာ (he is clever) and ဆရာများက သူ့ကို ချစ်ကြသည် (the teachers love him) is joined by postposition သောကြောင့် (as). The structure of the above sentence is Subj-Ada-CCS- Subj-Obj-Active. This is a correct sentence according to the Myanmar literature.

(a) NC [သူ/pron.person] # AC [လိမ္မာ/adj.dem] # CC [သောကြောင့်/cc.sent] # NC [ဆရာများ/n.objects] # PPC [က/ppm.subj] # NC [သူ့/pron.possessive] # PPC [ကို/ppm.obj] # VC [ချစ်ကြ/v.common] # SFC [သည်/sf]။

(b) Subj[သူ]#Ada[လိမ္မာ]#CCS[သောကြောင့်]#PSubj[ဆရာများ]#SubjP[က]#PObj [သူ့/pron.possessive] # ObjP [ကို/ppm.obj] # VC [ချစ်ကြ/v.common] # SFC [သည်/sf]။

(c)

| Sentence | [start] |
| <u>I-sent</u> CCS I-sent | [Sentence→I-sent CCS I-sent] |
| Subj Ada CCS <u>I-sent</u> | [I-sent→Subj Ada] |
| Subj Ada CCS <u>Subj</u> Obj Active | [I-sent→Subj Obj Active] |
| Subj Ada CCS PSubj SubjP <u>Obj</u> Active | [Subj → PSubj SubjP] |
| Subj Ada CCS PSubj SubjP PObj ObjP Active | [Obj → PObj ObjP] |

(d)
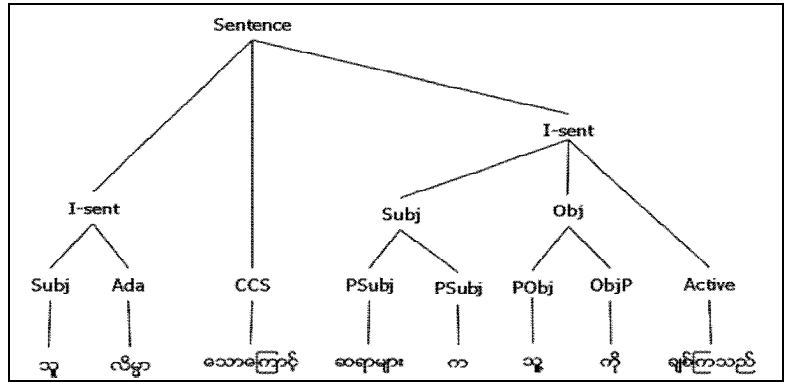

Figure 7. (a) The tagged and chunk complex sentence joined with conjunction (CCS)
(b) The function tagged sentence (c) Grammar derivation (d) The parse tree with function tags

## 10. EXPERIMENTAL RESULTS

In our corpus, all sentences can be further classified as two sets. One is simple sentence set, in which every sentence has no more than 15 words. The other is complex sentence set, in which every sentence has more than 15 words. There are 1600 simple sentences and 2300 complex sentences in the corpus.

For evaluation purpose, different numbers of sentences collecting from Myanmar textbooks of middle school and Myanmar historical books are used as a test set. There are about 2200 sentences in the test set. After implementation of the system using the grammar, it has been seen that the system can easily generates the parse tree for a sentence if the sentence structure satisfies the grammar rules. Our program tests only the sentence structure according to the grammar rules. So if the sentence structure satisfies the grammar rule, program recognizes the sentence as a correct sentence and generates a parse tree. Otherwise it gives output as an error.

Table 10 shows the overall performance for the proposed system. The proposed system yield 96.68% of precision, 93.05% of recall and 94.83% of f-measure for simple sentence. Performance comparisons between the various sentence types are shown in figure 8.

$$\text{Precision} = \frac{NumberOfCorrectSentences}{TotalNumberOfCorrectSentences} \times 100\%$$

$$\text{Recall} = \frac{NumberOfCorrectSentences}{NumberOfActualExistingCorrectSentences} \times 100\%$$

$$\text{F-Measure} = 2 * \frac{Precision * Recall}{Precision + Recall}$$

Table 10. Compared results of each sentence types

| Sentence Type | Actual | Recognized | Correct | Precision | Recall | F-Measure |
|---|---|---|---|---|---|---|
| Simple | 720 | 693 | 670 | 96.68% | 93.05% | 94.83% |
| Complex joined with CCP | 455 | 420 | 394 | 93.81% | 88.54% | 91.09% |
| Complex joined with CCA | 370 | 351 | 319 | 90.88% | 86.22% | 88.48% |
| Complex joined with CCS | 665 | 640 | 593 | 92.66% | 89.17% | 90.88% |

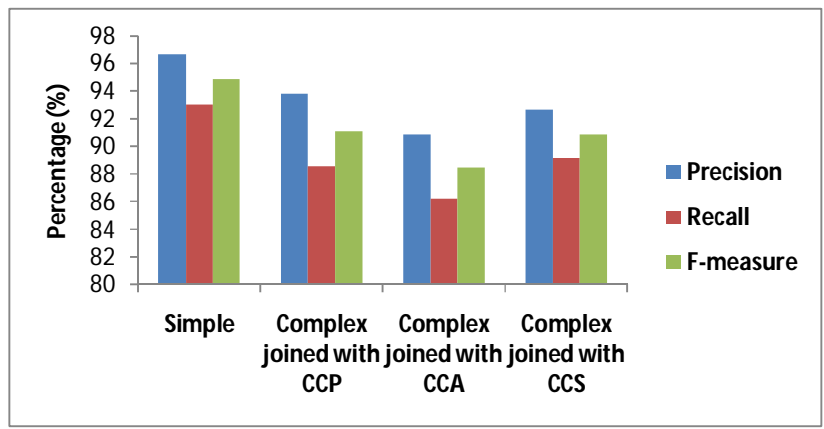

Figure 8. Performance Comparisons between the Various Sentence Types

## 11. CONCLUSION AND FUTURE WORK

In the task of assigning function tag, we chose Naïve Bayes model for its simplicity and user-friendliness. We apply context-free grammar for parsing because it is easier to maintain and can add new language features. The parse tree can be built by using function tags. As function tagging is a pre-processing step for parsing, the errors occurred in the task of function tagging affect the parse tree. The corpus may be balanced because Naïve Bayesian framework probability simply describes uncertainty. The corpus creation is time consuming. The corpus is the resource for the development of Myanmar to English translation system and we expect the corpus to be continually expanded in the future because the tested sentence can be added into the corpus.

In this work we have considered limited number of Myanmar sentences to construct the grammar rules. In future work we have to consider as many sentences as we can and some more tags for constructing the grammar rules because Myanmar language is a free-phrase-order language. Word position for one sentence may not be same in the other sentences. So we can not restrict the grammar rules for some limited number of sentences.

## REFERENCES

[1] John C. Henderson and Eric Brill. "Exploiting Diversity in Natural Language Processing: Combining Parsers".

[2] Blaheta, D (2003) "Function tagging". Ph.D. Dissertation, Brown University. Advisor-Eugene Charniak.

[3] Phyu Hnin Myint (2010) "Assigning automatically Part-of-Speech tags to build tagged corpus for Myanmar language", The Fifth Conference on Parallel Soft Computing, Yangon, Myanmar.

[4] Phyu Hnin Myint (2011) "Chunk Tagged Corpus Creation for Myanmar Language". In Proceedings of the ninth International Conference on Computer Applications, Yangon, Myanmar.

[5] Eugene Charniak (1997) "Statistical parsing with a context-free grammar and word statistics". In Proceedings of the Fourteenth National Conference on Artificial Intelligence, pages 598-603, Menlo Park.

[6] Blaheta, D., and Johnson, M (2000) "Assigning function tags to parsed text". In Proceedings of the 1st Annual Meeting of the North American Chapter of the Association for Computational Linguistics, 234–240.

[7] Mihai Lintean and Vasile Rus (2007) "Naive Bayes and Decision Trees for Function Tagging". In Proceedings of the International Conference of the Florida Artificial Intelligence Research Society (FLAIRS) 2007, Key West, FL, May (in press).

[8] Yong-uk Park and Hyuk-chul Kwon (2008) "Korean Syntactic Analysis using Dependency Rules and Segmentation ", Proceedings of the Seventh International Conference on Advanced Language Processing and Web Information Technology(ALPIT2008), Vol.7, pp.59-63, China, July 23-25.

[9] Mark-Jan Nederhof and Giorgio Satta (2002) "Parsing Non-Recursive Context-Free Grammars". In Proceedings of the 40th Annual Meeting of the Association for Computational Linguistics (ACL ANNUAL'02), July 7-12, Pages 112-119, Philadelphia, Pennsylvania, USA.

[10] Kyongho Min & William H. Wilson, "Are Efficient Natural Language Parsers Robust?" School of Computer Science & Engineering,University of New South Wales, Sydney NSW 2052 Australia

[11] Ethnologue (2005) Languages of the World, 15th Edition, Dallas, Tex.: SIL International. Online version: http://www.ethnologue.com/. Edited by Raymond G. Gordon, Jr.

[12] Myanmar Thudda (1986) vol. 1 to 5 in Bur-Myan, Text-book Committee, Basic Edu., Min. of Edu., Myanmar, ca.

[13] Shwe Pyi Soe, U (2010) မြန်မာဘာသာစကား Aspects of Myanmar Language

[14] Thaung Lwin, U (1978) နည်းသစ်မြန်မာသဒ္ဒါ

[15] Ko Lay, U (2003) မြန်မာသဒ္ဒါဖွဲ့စည်းပုံ Ph.D. Dissertation, Myanmar Department, University of Educaion.

[16] Win Win Thant (2010) "Naive Bayes for function tagging in Myanmar Language", The Fifth Conference on Parallel Soft Computing, Yangon, Myanmar, 2010.

[17] Leon Versteegen (1999) "The Simple Bayesian Classifier as a Classification Algorithm".

[18] Y. Tsuruoka and K. Tsujii (2005) "Chunk parsing revisited". In Proceedings of the Ninth International Workshop on Parsing Technologies. Vancouver, Canada.

[19] Michael Collins (1996) "A New Statistical Parser Based on Bigram Lexical Dependencies". In Proceedings of ACL-96, pp. 184–191.

**Authors**

**Win Win Thant is** a Ph.D research student. She received B.C.Sc (Bachelor of Computer Science) degree in 2003, B.C.Sc (Hons.) degree in 2004 and M.C.Sc (Master of Computer Science) degree in 2007. She is also an Assistant Lecturer of U.C.S.Y (University of Computer Studies, Yangon). She has published papers in International conferences and International Journals. Her research interests include Natural Language Processing, Artificial Intelligence and Machine Translation.

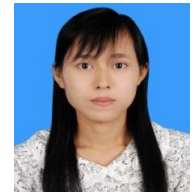

**Tin Myat Htwe** is an Associate Professor of U.C.S.Y. She obtained Ph.D degree of Information Technlogy from University of Computer Studies, Yangon. Her research interests include Natural Language Processing, Data Mining and Artificial Intelligence. She has published papers in International conferences and International Journals.

**Ni Lar Thein** is a Rector of U.C.S.Y. She obtained B.Sc. (Chem.), B.Sc. (Hons) and M.Sc. (Computer Science) from Yangon University and Ph.D. (Computer Engg.) from Nanyang Technological University, Singapore in 2003. Her research interests include Software Engineering, Artificial Intelligence and Natural Language Processing. She has published papers in International conferences and International Journals.

.